\documentclass[10pt,twocolumn,letterpaper]{article}

\usepackage{cvpr}
\usepackage{times}
\usepackage{epsfig}
\usepackage{graphicx}
\usepackage{amsmath}
\usepackage{amssymb}
\usepackage{url}
\usepackage{multirow}
\usepackage{epstopdf}

\usepackage[utf8]{inputenc} 
\usepackage[T1]{fontenc}    
\usepackage{url}            
\usepackage{booktabs}       
\usepackage{amsfonts}       
\usepackage{nicefrac}       
\usepackage{microtype}      
\usepackage{amsmath}
\usepackage{cleveref}
\usepackage{graphicx}
\usepackage[utf8]{inputenc}
\usepackage{caption}
\usepackage{subcaption}
\usepackage{bbm}
\usepackage{algorithm,algorithmic}
\usepackage[toc,page]{appendix}
\usepackage{authblk}

\usepackage{cleveref}
\crefname{section}{§}{§§}

\usepackage{adjustbox}
\usepackage{array}
\newcolumntype{R}[2]{%
    >{\adjustbox{angle=#1,lap=\width-(#2)}\bgroup}%
    l%
    <{\egroup}%
}

\usepackage[pagebackref=true,breaklinks=true,letterpaper=true,colorlinks,bookmarks=false]{hyperref}

\cvprfinalcopy

\begin{document}

\title{Learning Detection with Diverse Proposals}

\author{Samaneh Azadi$^1$, Jiashi Feng$^2$, and Trevor Darrell$^1$\\
$^1$University of California, Berkeley, $^2$National University of Singapore\\
{\hspace{-1.5cm}\tt\small \{sazadi,trevor\}@eecs.berkeley.edu} ${\qquad}$ {\tt\small{elefjia}@nus.edu.sg}}

\maketitle

\begin{abstract}

To predict a set of \emph{diverse} and \emph{informative} proposals with enriched representations,  this paper introduces a differentiable Determinantal Point Process (DPP) layer that is able to augment the object detection architectures. Most modern object detection architectures, such as Faster R-CNN, learn to localize objects by minimizing deviations from the ground-truth but ignore correlation ``between’’ multiple proposals and object categories. Non-Maximum Suppression (NMS) as a widely used proposal pruning scheme ignores label- and instance-level relations between object candidates resulting in multi-labeled detections. In the multi-class case, NMS selects boxes with the largest prediction scores ignoring the semantic relation between categories of potential election. In contrast, our trainable DPP layer, allowing for Learning Detection with Diverse Proposals (LDDP), considers both label-level contextual information and spatial layout relationships between proposals without increasing the number of parameters of the network, and thus improves location and category specifications of final detected bounding boxes substantially during \emph{both training and inference} schemes. Furthermore, we show that LDDP keeps it superiority over Faster R-CNN even if the number of proposals generated by LDPP is only $\sim$30\% as many as those for Faster R-CNN. \end{abstract}
  
\begin{figure}[t!]
\centering
\includegraphics[width=0.4\textwidth]{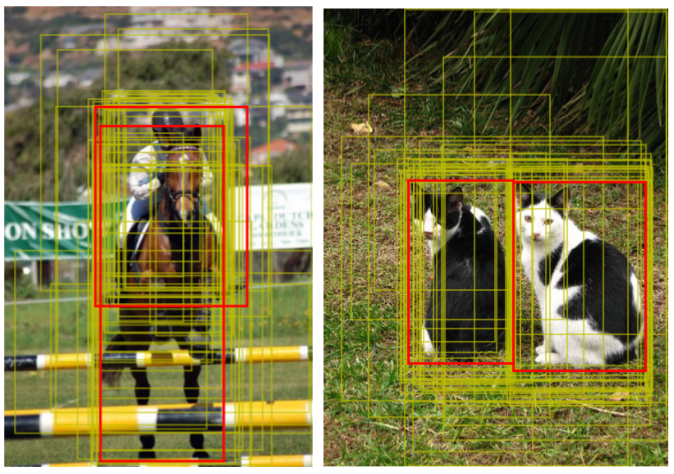}
\caption{Potential proposals as output from the region proposal network. There are many overlapping boxes on each object of the image whose prediction scores and location offsets are updated similarly in the Faster R-CNN network: the deviation of ``all'' proposals from their corresponding ground-truth should be minimized. However, the overlapping correlation between these proposals is ignored while training the model. We increase the probability of selecting the most representative boxes, shown in red, resulting in more diverse final detections. }
\label{fig:illustration}
\end{figure}
\section{Introduction}

Image classification~\cite{krizhevsky2012imagenet} and object detection~\cite{sermanet2013overfeat,girshick2014rich} have been improved significantly by development of deep convolutional networks~\cite{krizhevsky2012imagenet, lecun1989backpropagation}. However, object detection is still more challenging than image classification as it aims at both localizing and classifying objects. Accurate localization of objects in each image requires both well-processed ``candidate'' object locations and ``selected refined'' boxes with precise locations. Looking at the object detection problem as an extractive image summarization and representation task, the set of all predicted bounding boxes per image should be as informative and non-repetitive as possible.

The Region-based Convolutional Network methods such as Fast and Faster R-CNN~\cite{girshick2014rich,girshick2015fast,ren2015faster} proposed an efficient approach for object proposal classification and localization with a multi-task loss function during training. The training process in such methods contains a fine-tuning stage, which jointly optimizes a softmax classifier and a bounding-box regressor. Such a bounding box regressor tries to minimize the distance between the candidate object proposals with their corresponding ground-truth boxes for each category of objects. However, it does not consider relation ``between'' boxes in terms of location and context while learning a representation model. In this paper, we propose a new loss layer added to the other two softmax classifier and bounding-box regressor layers (all included in the multi-task loss for training the deep model) which formulates the discriminative contextual information as well as mutual relation between boxes into a Determinantal Point Process (DPP)~\cite{kulesza2011learning} loss function. This DPP loss finds a subset of diverse bounding boxes using the outputs of the other two loss functions (namely, the probability of each proposal to belong to each object category as well as the location information of the proposals) and will reinforce them in finding more accurate object instances in the end, as illustrated in Figure~\ref{fig:illustration}.  We employ our DPP loss to maximize the likelihood of an accurate selection given the pool of overlapping background and non-background boxes over multiple categories. 

Inference in state-of-the-art detection methods~\cite{wan2015end,girshick2015fast,ren2015faster, liu2015ssd} is generally based on Non-Maximum Suppression (NMS), which considers only the overlap between candidate boxes per class label and ignores their semantic relationship.  We propose a DPP inference scheme to select a set of non-repetitive high-quality boxes per image taking into account spatial layout, category-level analogy between proposals, as well as their quality score obtained from deep trained model. We call our proposed model as ``Learning Detection with Diverse Proposals Network -- LDDP-Net''.      

Our proposed loss function for representation enhancement and more accurate inference can be applied on any deep network architecture for object detection. In our experiments below we focus on the Faster R-CNN model to show the significant performance improvement added by our DPP model. We demonstrate the effect of our proposed DPP loss layer in accurate object localization during training as well as inference on the benchmark detection data sets PASCAL VOC and MS COCO based on average precision and average recall detection metrics. 

To sum up, we make following contributions in this work:
\begin{itemize}
\item We propose to explicitly pursue diversity on generated object proposals and introduce the strategy of learning detection with diverse proposals.
\item We introduce a DPP layer that is able to maximize diversity favorably in an end-to-end trainable way. Besides it is compatible with many existing state-of-the-art detection architectures and thus able to augment them effectively.
\item Experiments on Pascal VOC and MS COCO data sets clearly demonstrate the superiority of diverse proposals and effectiveness of our proposed method on producing diverse detections.
\end{itemize}

LDDP Code is available at \url{https://github.com/azadis/LDDP}.

 \section{End-to-End LDDP Model}
 \label{sec:LDDP-net}
Faster R-CNN~\cite{ren2015faster} as a unified deep convolutional framework for generating and refining region proposals alternates between fine-tuning for proposals using a fully convolutional Region Proposal Network (RPN) and fine-tuning for object detection by Fast R-CNN model. Keeping the object proposals generated from RPN fixed, they will be mapped into convolutional features through several convolutional and max-pooling layers. An RoI pooling layer converts the features inside each region of interest (RoI) into a fixed-length feature vector afterwards, which will be fed into a sequence of fully connected layers. 

The loss function on the top layer of detection model is a multi-task loss dealing with both classification and localization of object proposals: the softmax loss layer outputs a discrete probability distribution over the $K+1$ categories of objects in addition to the background for each object proposal, and the bounding box regressor layer determines location offsets per object proposal for all categories.

 Applying a diversity-ignited model can reinforce the network to limit the boxes around each object while they have minimum overlap with other object bounding boxes in the image. It will also ``select'' boxes in such a way to make their collection as informative as possible given the minimum possible number of proposals. We define such a model through a DPP loss layer added to the other two loss functions introduced in the Faster R-CNN architecture. 

On the other hand, inference in Region-based CNN models as well as other state-of-the-art networks is done through NMS which selects boxes with highest detection scores for each category. Giving a priority to the detection scores, it might finally end up in selecting overlapping detection boxes and miss the best possible set of non-overlapping ones with acceptable score. Besides, NMS neglects the semantic relations between categories as its selection is done category-wisely. We address all such problems through a probabilistic DPP inference scheme which jointly considers all spatial layout, contextual information, and semantic similarities between object candidates and selects the best probable ones. In the following section, we define our learn-able DPP loss layer, show how to back-propagate through this layer while training the model (summarized in Alg.~\ref{alg-2}), and then clarify how to infer the best non-overlapping boxes from the predicted proposals.

\begin{algorithm}
\caption{LDDP Learning}
\label{alg-2}
\begin{algorithmic}
\STATE \textbf{Input} Set $X:\{i: i \in \text{mini-batch}\}$,\\
$\{b_i^c,t_i^c: \quad\forall i \in X, c=0,\cdots,K\}$: box probability and offset values, \\
  \textbf{Output} Loss $\mathcal{L}$, $\partial \mathcal{L}/\partial b_i^c \quad \forall i \in X$\\
  \STATE $X_s \leftarrow $ subset  of $X$ including non-background proposals with high overlap with gt boxes\\
  \STATE $\Phi_i\leftarrow \text{IoU}_{i,gt^i} \quad \forall i \in X_s $,\\
  \STATE $Y\leftarrow$ Apply Alg.~\ref{alg-1} on $X_s$,\\
 \STATE  $B\leftarrow$ background proposals as defined in $\S$ \ref{sec-phi},\\
 \STATE $\Phi_i \leftarrow \text{Eq}.~\eqref{phi-p},~\eqref{phi-q} \quad \forall i \in X$,\\
\STATE $\mathcal{L} \leftarrow \text{Eq}.~\eqref{log-like},~\eqref{DPP}$,\\
\STATE $\partial \mathcal{L}/\partial b_i^c \leftarrow \text{Eq}.~\eqref{grad-p},~\eqref{grad-q} \quad \forall i \in X, c \in \{0,\cdots,K\}$
  \RETURN $\mathcal{L}, \partial \mathcal{L}/\partial b_i^c $
\end{algorithmic}
\end{algorithm}

\subsection{Learning with Diverse Proposals}
\label{sec-learn}
Determinantal Point Processes (DPPs) are natural models for diverse subset selection problems~\cite{kulesza2011learning}. In the selection problem, there is a trade-off between two influential metrics: The selected subset of items, or in other words their summary, should be ``representative'' and cover significant amount of information from the whole set. Besides, the information should be passed ``efficiently'' through this selection; the selection should be diverse and non-repetitive. We briefly explain a determinantal point process here and refer the readers to~\cite{kulesza2011learning} for an in-depth discussion. 
\subsubsection{Determinantal Point Process}
A point process $\mathcal{P}$ on a discrete set $X=\{x_1,\cdots, x_N\}$ is a probability measure on the set of all subsets of $X$. $\mathcal{P}$ is called a determinantal point process (DPP) if:
\begin{eqnarray*}
P_L(\mathbf{Y} = Y) = \frac{\det(L_Y)}{\det(L+I)}
\end{eqnarray*}
where $I$ is an identity matrix, $L$ is the kernel matrix, and $Y$ is a random subset of $X$ . The positive semi-definite kernel matrix $L$ indexed by elements of $Y$ models diversity among items: the marginal probability of inclusion of each single item is proportional to the diagonal values of the kernel matrix $L$, while correlation between each pair of items is proportional to the off-diagonal values of the kernel. As a result, subsets with higher diversity measured by the kernel have higher likelihood.

In the object detection scenario, items are indeed the set of proposals in the image produced by the region proposal network. Given the bounding box probability scores of softmax loss layer and location offsets from the bounding box regressor layer for the given set of proposals for image $i$, $X^i$, we seek for a precise and diverse set of boxes, $Y^i$. In other words, the probability of selecting informative non-redundant boxes given the whole set of background and non-background proposals should be maximized during training. Simultaneuosly, the probability of selecting background boxes, denoted by $B^i$, should be minimized.  We employ a learn-able determinantal point process layer by maximizing the log-likelihood of the training set \emph{and} minimizing the log-likelihood of background proposals election:
\begin{eqnarray}
\mathcal{L}(\alpha) =&& \log \prod_i \frac{P_{\alpha}(Y^i|X^i)}{P_{\alpha}(B^i|X^i)} \nonumber\\
=&& \sum_i \big[ \log P_{\alpha}(Y^i|X^i) -  \log P_{\alpha}(B^i|X^i)\big] \label{log-like}
\label{eq:loss}
\end{eqnarray} where $\alpha$ refers to the parameters of the deep network.

For simplicity, we assume that number of images per iteration is one and thus, remove index $i$ from our notations. We follow the same mini-batch setting as in Faster R-CNN network~\cite{ren2015faster} where $m$ is the number of object proposals in each iteration or the size of mini-batch.

 Given a list of object proposals as output of the RPN network, $X$, a posterior probability $P_{\alpha} (Y|X)$ modeled as a determinantal point process would imply which boxes should be selected with a high probability:
\begin{eqnarray}
\label{DPP}
P_{\alpha}(\mathbf{Y}=Y|X) &=& \frac{\det(L_Y)}{\det(L+I)} , \nonumber  \\
L_{i,j} &=& \Phi_i^{1/2}S_{ij}\Phi_j^{1/2}, \nonumber \\ 
S_{ij} &=&\text{IoU}_{ij} \times \text{sim}_{ij},\nonumber\\
\text{IoU}_{ij} &=& \frac{A_i \cap A_j}{A_i \cup A_j}, \nonumber\\ \text{sim}_{ij} &=& \frac{2IC(l_{cs}(C_i,C_j))}{IC(C_i)+IC(C_j)}. 
\end{eqnarray}
The above distribution model considers relation among different proposals (indexed by $i,j$) through their similarity encoded by $S_{ij}$ (which is the product of their spatial overlap rate and category similarity) as well as their individual quality score $\Phi_i$. We now proceed to explain each quantity in the distribution model.

\begin{algorithm}[t]
\caption{LDDP Inference}
\label{alg-1}
\begin{algorithmic}
\STATE \textbf{Input} Set $X$: Set of proposals and their prediction scores and box offsets,\\
	$\mathcal{T}:$ fixed threshold\\
  \textbf{Output} Set $Y$: Non-overlapping representative proposals\\
  \STATE $Y\leftarrow \emptyset$, $Y' \leftarrow X$,\\
  \WHILE{\text{len}$(Y) < \#$ \text{Dets} \AND $Y'\neq \emptyset$}
 \STATE \text{cost}$(i) \leftarrow \max_j S_{ij}$\\
 \STATE $k \leftarrow \arg \max_{i \in Y'} P_{\alpha}(Y \cup i|X)$
  \IF {\text{cost}$(k)<\mathcal{T}$ }
  \STATE {$Y \leftarrow Y \cup k$}\\
  \ENDIF\\
  $Y' \leftarrow Y' \backslash k$
  \ENDWHILE \\
  \RETURN $Y$
\end{algorithmic}
\end{algorithm}

We set $S = S + \epsilon I$ with a small $\epsilon>0$  to make sure that the ensemble matrix $L$ is positive semi-definite, which is important for a proper DPP model definition. Here, $\det (L+I)$ is a normalizing factor, and $\text{IoU}_{ij}$ is the Intersection-over-Union associated with each pair of proposals $i$ and $j$, where $A_i$ is the area covered by the proposal $i$. Motivated by ~\cite{lin1998information, mrowca2015spatial}, we consider the semantic relation among proposals, $\text{sim}_{ij}$, as the semantic similarity between the labels of each pair of proposals $(i,j)$. Here $l_{cs}(C_i,C_j)$ refers to the lowest common subsumer of the category labels $C_i$ and $C_j$ in the WordNet hierarchy. The information content of class $C$ is computed as $IC(C) = -\log P(C)$ where $P(C)$ is the probability of occurrence of an instance with label $C$ among all images. Posterior probability of selecting background boxes, $P_{\alpha}(B|X)$, follows the same determinantal point process formulation as in Eq.~\eqref{DPP}. The difference between these two posterior probabilities lies in how we measure quality of proposals $\Phi$.    
\subsubsection{Model Description}
\label{sec-phi}
In general, the classification score over $K+1$ categories as well as overlap with the bounding box target can be used to define the quality score $\Phi_i$ for proposal $i$. The classification scores are computed in the fully-connected layer before ``softmax'' layer in the Faster R-CNN architecture, and location of each bounding box is given by the inner product layer feeding into the ``bounding box regressor''.  The exact definition of $\Phi_i$ for different proposals in the two terms of log-likelihood function Eq.~\eqref{eq:loss} varies according to the general goal of increasing scores of high-quality boxes in their ground-truth label and increasing scores of background boxes in the background category. 

One should note that $Y$ is an ideal extractive summary of input which is the list of proposals in each \textit{mini-batch}, $X$ (e.g. $m=128$)~\cite{kulesza2011learning}. Thus, we apply a maximum a posteriori (MAP) DPP inference, Alg.~\ref{alg-1}, in each iteration of the training algorithm to determine the set of representative boxes $Y$ from the set of $m$ proposals. If ground-truth boxes exist among the proposals in each mini-batch, they will be automatically selected as the set $Y$ through MAP. However, if they don't exist among the proposals, MAP will select the best summarizing ones. To make sure that selected boxes, $Y$, are accurate and close to ground-truth boxes, we only select from a ``subset'' of proposals in the mini-batch with high overlap with their bounding box targets. Also, we define the quality of boxes only based on their overlap with their bounding box target in this step of applying MAP, as summarized in Alg.~\ref{alg-2}. Thus, by selecting $Y$ as the set of best representations of ground-truth boxes in $X$, maximizing $P(Y|X)$ results in maximizing selection of ground-truth boxes, which corresponds to maximizing the probability of training data. 

On the other hand, to specify the set of more probable background proposals as $B$, given the set of proposals in each mini-batch $X$, we define $B$ as the set of all proposals in $X-Y$ except those that have high overlap with ground-truth \textit{and} their associated predicted label matches their ground-truth label. As mentioned before, the goal here is to \emph{minimize} the probability of selecting ``background proposals'' as the representative boxes in each image. 

To complete the DPP model description, we define the quality scores of the proposals as follows.

For the first term $\log P_{\alpha}(Y|X)$, we define the quality of boxes as:
\begin{eqnarray}
\label{phi-p}
\Phi_{i} = \begin{cases} 
\text{IoU}_{i,gt^i} \times \exp\{W_{gt}^Tf_i\}\label{phi-p} ,& \text{if } i \in Y\\
 \text{IoU}_{i,gt^i} \times \sum_{c \neq 0}\exp\{W_{c}^Tf_i\} &\text{if } i \not \in Y
\end{cases}
\end{eqnarray}
where $W_{c}^Tf_i$ is the output of the inner product layer before the softmax loss layer, $W_{c}$ is the weight vector learned for category $c$ and $f_i$ is the fc7 feature learned for proposal $i$. Moreover, $W_{gt}$ denotes the weight vector for the corresponding ground-truth label of proposal $i$, and $c=0$ shows the background category. Note that the goal is to increase the score of boxes in $Y$ in their ground-truth label and the score of other boxes in the back-ground category. Derivatives of log-likelihood function with respect to $W_c^Tf_i$ in Eq.~\eqref{grad-p},~\eqref{grad-q} clarifies all above definitions.

The second term $\log P_{\alpha} (B|X)$ is designed for minimizing the probability of selection of background boxes which results in a boost in the scores of such proposals in their background category. We thus define the quality of proposals for this term as:
\begin{eqnarray}
\label{phi-q}
\Phi_{i} = \begin{cases} 
\text{IoU}_{i,gt^i} \times \sum_{c \neq 0}\exp\{W_{c}^Tf_i\}\label{phi-q} ,& \text{if } i \in B\\
\text{IoU}_{i,gt^i} \times \exp\{W_{gt}^Tf_i\} &\text{if } i \not \in B
\end{cases}
\end{eqnarray}
The effect of involving $\text{IoU}_{i,{gt}^i}$ in quality scores during training appears in computing the gradient, Eq.~\eqref{grad-p},\eqref{grad-q}, where a larger gradient would be passed for boxes with higher overlap with their bbox target. It means more accurate boxes will move toward being selected (achieve higher softmax prediction scores) faster than others.

To avoid degrading the relatively accurate bounding boxes which are not selected in $Y$, in Eq.~\eqref{eq:loss}, during the learning process, we exclude from $X$ all the boxes that have high overlap with boxes in $Y$ and their label matches their ground-truth category (in both $P_{\alpha}(Y^i|X^i)$, $P_{\alpha}(B^i|X^i)$).

\subsubsection{LDDP Back-Propagation}
We modify the negative log-likelihood function in Eq.~\eqref{log-like} such that the two terms get balanced according to the number of selected proposals in $Y, B$. This DPP loss function, depends on the inputs of both softmax loss and bounding box regressor in the deep network. Since the ensemble matrix $L$, Eq.~\eqref{DPP}, and the consequent log-likelihood function in Eq.~\eqref{log-like} are functions of the parameters of deep network, we incorporate the gradient of our DPP loss layer into the back-propagation scheme of the detection network. We assume the location coordinates of proposals are fixed and only consider the outputs of the fully-connected layer before softmax loss as the parameters of our loss layer.

Based on definitions of $\Phi_i$ in Eq.~\eqref{phi-p} and model presented in Eq.~\eqref{DPP}, the gradient of $p_1 =\log P_{\alpha} (Y|X)$ with respect to $b_i^c = W_{c}^Tf_i$, the output of the inner product layer, would be as follows: 
\begin{eqnarray}
\label{grad-p}
\frac{-\partial \log p_1}{\partial b_i^c} = \begin{cases} 
K_{ii}-1 ,& \text{if } i \in Y, c=\text{gt}\\
 \frac{ K_{ii} \exp\{b_i^c\}}{\sum_{c'\neq 0} \exp\{b_i^{c'}\}}, &\text{if } i \not \in Y,  c\neq 0\\
0 & \text{otherwise}
\end{cases}
\end{eqnarray} where $K_{ii}= L_{ii}/\det(L+I)$. Therefore, minimizing the negative log-likelihood increases the scores of representative boxes in their ground-truth label and background boxes in background label. Similarly, according to the defined $\Phi_i$'s in Eq.~\eqref{phi-q}, the gradient of $p_2 =\log P_{\alpha}(B|X)$ w.r.t $b_i^c$ is:

\begin{eqnarray}
\label{grad-q}
\frac{\partial \log p_2}{\partial b_i^c} = \begin{cases} 
-K_{ii} ,& \text{if } i \not \in B,  c=\text{gt}\\
 \frac{-(K_{ii}-1) \exp\{b_i^c\} }{\sum_{c'\neq 0} \exp\{b_i^{c'}\}}, &\text{if } i \in B,  c\neq 0\\
0 & \text{otherwise}
\end{cases}
\end{eqnarray}

Consequently, the gradient of the above log-likelihood function with respect to $b_i^c$ will be added to the gradient of the other two loss functions in the \emph{backward} pass while end-to-end training of parameters of the network. The proof for the above gradient derivations is provided in Appendix~\ref{sec:proof}.

\subsection{Inference with Diverse Proposals}
Given the prediction scores and bounding box offsets from the learned network, we model the selection problem for unseen images as a maximum a posteriori (MAP) inference scheme where the probability of inclusion of each candidate box depends on the determinant of a kernel matrix. We define this kernel or similarity matrix such that it captures all spatial and contextual information between boxes all at once. We use the same kernel matrix $L$ as in Eq.~\eqref{DPP} where $X$ is the list of all candidate proposals per image over all $K$ non-background categories with a score above an specific threshold (e.g. 0.05).

We capture quality of boxes, $\Phi$, by their per class prediction scores:
\begin{eqnarray*}
\Phi_i = \frac{\exp\{W_c^Tf_i\}}{\sum_{c'} \exp\{W_{c'}^Tf_i\}} \quad \text{for } c\in \{0,\cdots, K\}
\end{eqnarray*}
Similarly, we employ the spatial information by $\text{IoU}$ and the semantic similarity between box labels by $\text{sim}_{ij}$ as shown in Eq.~\eqref{DPP}.

Thus, our kernel definition allows selection of a set of candidate boxes which have minimum possible overlapping as well as highest detection scores. In other words, the boxes with less location- and label-level similarity and higher detection scores would be more probable to be selected. To figure out which boxes should be selected, similar to ~\cite{kulesza2011learning}, we use a greedy optimization algorithm, Alg.~\ref{alg-1}, which keeps the box with the highest probability found by Eq.~\eqref{DPP} at each iteration.

\begin{table*}[!t]
\centering
\caption{VOC2007 test detection average precision($\%$) (trained on VOC2007 trainval) at IoU thresholds o.5 and 0.7. All methods use ZF deep convolutional network. Each pair $(x,y)$ indicates method $x$ used for learning and $y$ for inference. In both tables, the two top rows use NMS and the two bottom rows show LDDP used for inference. Here ``FrRCNN'' refers to ``Faster RCNN''} 
\label{voc-tab}
\footnotesize
\begin{tabular}{|p{0.05cm}|p{2.05cm}|p{0.27cm}p{0.27cm}p{0.27cm}p{0.27cm}p{0.27cm}p{0.27cm}p{0.27cm}p{0.27cm}p{0.27cm}p{0.27cm}p{0.27cm}p{0.27cm}p{0.27cm}p{0.37cm}p{0.27cm}p{0.27cm}p{0.27cm}p{0.27cm}p{0.27cm}p{0.27cm}|p{0.36cm}|}
\hline	
&Method            & aero          & bike          & bird          & boat          & bottle        & bus           & car           & cat         & chair         & cow           &  table         & dog           & horse         & mbike        & persn        & plant         & sheep         & sofa          & train         & tv        & mAP   \\
\hline
\parbox[t]{.05cm}{\multirow{4}{*}{\rotatebox[origin=c]{90}{@ IoU 0.5 }}}  &(FrRCNN, NMS)  & 63.7          & 70.1          & 55.5          & 45.4          & 37.1          & 66.3          & 75.5          & 71.5          & 39.3        & 66.3          & 60.2          & 61.5          & 76.7          & 69.7          & \textbf{71.5} & 34.2          & 53.3          & 55.9          & 69.9          & \textbf{65.3} & 60.45 \\
&(LDDP, NMS)       & 65.6          & 72.7          & 56.3          & 44.6          & 36.9          & 67.8          & 75.5          & 73            & 39.4        & 63.5          & \textbf{61.8} & 65.7          & 74.9          & 70.9          & 71.2          & 37.4          & 55.7          & 55.8          & 71.4          & 62.6          & 61.14 \\
\cline{2-23}
\cline{2-23}
&(FrRCNN, LDDP) & \textbf{66.1} & 70.4          & 56.5          & \textbf{45.8}          & \textbf{37.1} & 67.8          & \textbf{75.5} & 71.1          & \textbf{40} & \textbf{68.9} & 59.9          & 65.8          & \textbf{78.7} & \textbf{71.6} & 71.4          & 34            & \textbf{59.7} & 56.4          & \textbf{72.3} & 64.9          & 61.7  \\
&(LDDP, LDDP)      & 66            & \textbf{73.2} & \textbf{56.9} & 45.6 & 37            & \textbf{70.1} & 75.4          & \textbf{74.9} & 39.4        & 66.6          & 61.5          & \textbf{68.5} & 77.4          & 71.5          & 71            & \textbf{37.7} & 58.8          & \textbf{56.6} & 71.9          & 64.1           & \textbf{62.21}\\
\hline
\multicolumn{22}{c}{ }\\
 \hline
 \parbox[t]{0.05cm}{\multirow{4}{*}{\rotatebox[origin=c]{90}{@ IoU 0.7 }}}  
 &(FrRCNN, NMS) & 36.0 & \textbf{45.8} & 25.5 & 18.0   & 14.8 & 46.6 & 55.2 & 41.9 & 17.1 & 30.8 & 38.8 & 30.9 & 48.1 & 45.1 & 35.2 & 13.8 & 30.7 & 30.3 & 43.8 & 44.2 & 34.6 \\
&(LDDP, NMS)        & 37.6 & 44.8 & 22.8 & 19.7 & 16.2 & \textbf{50.8} & 56.5 & \textbf{45.2} & 19.2 & 36.7 & 39.2 & 35.6 & 48.9 & 45.4 & 36.9 & 13.7 & 35.5 & 35.1 & 42.6 & 40.8 & 36.2 \\
\cline{2-23}
\cline{2-23}
 &(FrRCNN, LDDP) & 36.4          & 45.7 & \textbf{26.9} & 19.3          & 15.3          & 47.7          & 55.0            & 41.1          & 17.2          & 31.5          & 38.7          & 34.4          & 48.1          & \textbf{49.1} & 35.3        & 13.9          & 31.1          & 32.3          & 44.1          & \textbf{45.6} & 35.4          \\
&(LDDP, LDDP)        & \textbf{38.2} & 45.2          & 25.2          & \textbf{20.9} & \textbf{16.4} & 50.7 & \textbf{56.8} & 45.1 & \textbf{19.5} & \textbf{37.2} & \textbf{39.8} & \textbf{35.7} & \textbf{49.6} & 46.3          & \textbf{37} & \textbf{14.5} & \textbf{36.4} & \textbf{35.2} & \textbf{44.7} & 40.9          & \textbf{36.8}\\
\hline
\end{tabular}
\end{table*}

\section{Related Work}
Several works~\cite{desai2011discriminative, mrowca2015spatial, lee2016individualness} have proposed a replacement for the conventional non-maximum suppression to optimally select among detection proposals. Desai et al.~\cite{desai2011discriminative} proposed a unified model for multi-class object detection that both learns optimal NMS parameters and infers boxes by capturing different structured contextual interactions between object and their categories. In contrast to~\cite{desai2011discriminative}, the contextual information used in our model captures fixed label-level semantic similarities based on WordNet hierarchy as well as learned proposal probability scores. We also use IoU to capture spatial layout within a similarity kernel, while there is no strong notion of quality or overlap among boxes in~\cite{desai2011discriminative} and only a thresholded value is used in a (0/1) loss function. Unlike this method, the learnable deep features in our full end-to-end framework improve bounding box locations and category specifications through our proposed differentiable loss function.

Mrowca \textit{et al}.~\cite{mrowca2015spatial} proposed a large-scale multi-class affinity propagation clustering (MAPC)~\cite{frey2007clustering} to improve both the localization and categorization of selected detected proposals, which simultaneously optimizes across all categories and all proposed locations in the image. Similar to our semantic similarity metric, they use WordNet relationships to capture highly related fine-grained categories in a large-scale detection setting. Lee \textit{et al}.~\cite{lee2016individualness} use individualness to measure quality and similarity scores in a determinantal point process inference scheme focusing on the ``binary'' pedestrian detection problem. However, these methods are only applied for the inference paradigm and can neither improve proposal representations nor impose diversity among object bounding boxes while training the model. 

\begin{table*}[!t]
\centering
\caption{VOC2007 test detection average precision($\%$) (trained on VOC2012 trainval) at IoU thresholds o.5. All methods use ZF deep convolutional network. Each pair $(x,y)$ indicates method $x$ used for learning and $y$ for inference. In both tables, the two top rows use NMS and the two bottom rows show LDDP used for inference. Here ``FrRCNN'' refers to ``Faster RCNN''} 
\label{voc-tab-2012}
\footnotesize
\begin{tabular}{|p{0.05cm}|p{2.05cm}|p{0.27cm}p{0.27cm}p{0.27cm}p{0.27cm}p{0.27cm}p{0.27cm}p{0.27cm}p{0.27cm}p{0.27cm}p{0.27cm}p{0.27cm}p{0.27cm}p{0.27cm}p{0.37cm}p{0.27cm}p{0.27cm}p{0.27cm}p{0.27cm}p{0.27cm}p{0.27cm}|p{0.36cm}|}
\hline	
&Method            & aero          & bike          & bird          & boat          & bottle        & bus           & car           & cat         & chair         & cow           &  table         & dog           & horse         & mbike        & persn        & plant         & sheep         & sofa          & train         & tv        & mAP   \\
\hline

\parbox[t]{.05cm}{\multirow{4}{*}{\rotatebox[origin=c]{90}{@ IoU 0.5 }}}  &(FrRCNN, NMS)  & 66.3          & 68.2          & 56.4          & 47.4          & 37.7          & 65.0          & 73.8          & 76.2         & 37.5       & 63.3          & 57.5         &68.7          & 73.5          & 70.0          & 70.3 & 35.1          & 64.3          & 57.5          & 66.8         & 63.1 & 60.9 \\
&(LDDP, NMS)       & \textbf{67.5}         & 70.6         & 57.1          & 47.6          & \textbf{40.4}          & 66.9          & 73.1         & 73.9            & 40.6      & 65.1          & 58.2 & 67.9          & 72.4         & 70.4          & \textbf{70.6}          & 35.8          & 63.9          & 58.2          & 70.9          & 64.7          & 61.8 \\
\cline{2-23}
\cline{2-23}

&(FrRCNN, LDDP) & 66.9 & 68.8          & 57.4          & \textbf{48.5}          & 37.2 & 67.9          & 73.8 & \textbf{78.0}         & 38.6 & 66.2 & \textbf{58.5}          & \textbf{71.1}         & \textbf{76.4} & \textbf{71.7} & 70.3          & 35.8            & 66.1 & 58.7         & 68.6 & 63.4          & 62.2  \\

&(LDDP, LDDP)      & 67.4            & \textbf{72.5} & \textbf{57.6} & 48.3 & 40.3            & \textbf{68.9} & \textbf{74.8}          & 76.2 & \textbf{41.1}        &\textbf{68.0}          & 58.2          & 70.8 & 74.8          & 71.1          & 70.5            & \textbf{36.2} & \textbf{66.8}          & \textbf{59.2} & \textbf{72.1}          & \textbf{64.7}           & \textbf{63.0}\\
\hline
\end{tabular}
\end{table*}

\section{Experiments}
We demonstrate the significant improvement obtained from our model on the representation of detected bounding boxes via a set of quantitative and qualitative experiments on Pascal VOC2007 and MS COCO benchmark data sets as explained in the following sections. We use the \textrm{caffe} deep learning framework~\cite{jia2014caffe} in all our experiments. We use one image per iteration with  mini-batch of 128 proposals. We replaced the semantic similarity matrix in Eq.~\eqref{DPP} with its fourth power during LDDP inference as we observed improvement on detection performance on validation sets.

Our baseline in all experiments is the state-of-the-art object detection Faster R-CNN approach used as our training model. Moreover, final detections are inferred by the NMS scheme applied on top of the deep trained network, denoted as our inference baseline. We use two different NMS IoU threshold values for within class and across class suppressions: First, proposals labeled similarly are suppressed by an IoU threshold resulting in less overlapping within-class detections. Afterwards, we apply a second IoU threshold to suppress boxes across different categories. We do a grid-search on these two thresholds to find the best combination.

\subsection{Experiments on Pascal VOC2007}
\begin{figure}[t!]
\centering
\includegraphics[width=0.48\textwidth]{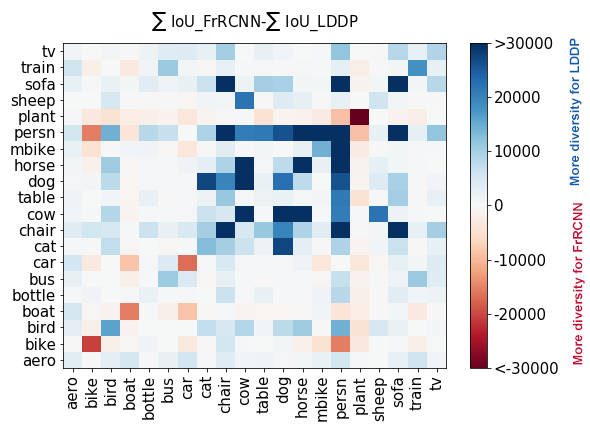}
\caption{Difference between sum of IoUs between proposals in different categories of VOC2007 data set generated by Faster R-CNN and LDDP.}
\vspace{-0.3cm}
\label{fig:diversity}
\end{figure}

We evaluate the performance of our proposed LDPP model on Pascal VOC2007 data set with 5k trainval images and 5k test images categorized with 20 different labels. We use the ImageNet pre-trained fast Zeiler and Fergus (ZF) model~\cite{zeiler2014visualizing} with 5 convolutional layers and 3 fully-connected layers and add our DPP loss layer to the other two existing loss functions in the multi-task loss. We evaluate our results based on the mean Average Precision (mAP) metric for object detection for both training and inference schemes. Results in Table~\ref{voc-tab} show the significant improvement made by our proposed LDDP model over the state-of-the-art Faster R-CNN approach on per-category average precision scores and their corresponding mean average precision (mAP), both in training and inference steps.
\begin{table*}[]
\centering
\caption{MS COCO val detection average precision and average recall($\%$) (trained on COCO train set). All methods use \text{VGG\_CNN\_M\_1024} deep convolutional network.  Each pair $(x,y)$ indicates method $x$ used for learning and $y$ for inference. The two top rows use NMS and the two bottom rows show LDDP used for inference.}
\label{coco-tab}
\begin{tabular}{|l|lll|lll|lll|lll|}
\hline
\multirow{2}{*}{Method} & \multicolumn{3}{l|}{Avg Precision @ IoU:} & \multicolumn{3}{l|}{Avg Precision @ Area:} & \multicolumn{3}{l|}{Avg Recall, \#Dets:} & \multicolumn{3}{l|}{Avg Recall @ Area:} \\ 
                        & 0.5-0.95      & 0.5         & 0.75        & S            & M            & L            & 1            & 10          & 100         & S           & M           & L           \\ \hline
(Faster RCNN, NMS)      & 15.0        & 31.5      & 12.7      & 3.6       & 15.1       & 23.8       & 16.4       & 23.2      & 23.6      & 6.0      & 24.2      & 38.9      \\ 
(LDDP, NMS)             & 15.2        & 31.5      & 13.1      & 3.4        & 15.5       & 24.1       & 16.6       & 23.7      & 24.1      & 6.1      & 25.0      & 39.4      \\ \hline
\hline

(Faster RCNN, LDPP)             & 15.3        & \textbf{32.3}     & 12.9      & \textbf{3.7  }      & 15.5       & 24.5      & 17.3       & 24.9      & 25.4      & 6.7      & 26.5      & 42.5      \\ 
(LDDP, LDPP)             & \textbf{15.5}        & 32.2      & \textbf{13.4 }     & 3.5        & \textbf{15.8}       & \textbf{24.7}      & \textbf{17.4}       & \textbf{25.4}      & \textbf{26.0}      & \textbf{6.8}      & \textbf{27.3 }     & \textbf{43.2}   \\ \hline
\end{tabular}
\end{table*}
In addition, to compare diversity among proposals generated by the two training models, Faster R-CNN and our proposed LDDP, we add up the intersection-over-union values for proposals among the $K=20$ categories with prediction scores above $0.05$ and compare the results in Figure~\ref{fig:diversity}. The total number of proposals generated by the two approaches are similar in this experiment. As expected, our LDDP model generates less overlapping proposals in most of the categories. For instance, the number of overlapping boxes identified as ``horse'' and ``dog'' is much larger in the Faster R-CNN approach than our LDDP model. This behavior is observed for other pairs of categories such as (dog, cat), (dog, cow), (person, sofa), (person, horse), (sofa, chair), etc revealing detection of diverse proposals through our proposed DPP loss layer.

\subsection{Experiments on Pascal VOC2012}

We repeated similar experimental setup on Pascal VOC2012 data set containing 11.5K trainval images~\cite{pascal-voc-2012} and use the Imagenet pretrained ZF network~\cite{zeiler2014visualizing} to train the end-to-end model and tested the results on Pascal VOC2007~\cite{pascal-voc-2007} data set.
The results comparing our LDPP model used either in training, inference, or both (end-to-end LDDP) with the state-of-the-art Faster R-CNN detection network~\cite{ren2015faster}, in Table~\ref{voc-tab-2012}, show the out-performance of our proposed model. 

\subsection{Experiments on Microsoft COCO}
Next, we demonstrate results on Microsoft COCO data set containing 80 different object categories. We train our model on the 80K training set and evaluate on the 40K validation set and use the \text{VGG\_CNN\_M\_1024} deep network architecture pre-trained on ImageNet data set~\cite{chatfield2014return, girshick2015fast}. In addition to the Average Precision metric (AP) at multiple IoU threshold values and size of object proposals, we evaluate our proposed model based on Average Recall (AR) when allowing a different range for maximum number of detections or size of proposals. In all cases represented in Table~\ref{coco-tab}, our proposed LDDP model outperforms the state-of-the-art Faster R-CNN network on COCO detection.     

As an ablation study, we used different powers of the semantic similarity matrix during inference and evaluated the detection performance on the minival subset of MS COCO data set with 5K images. As shown in Table~\ref{coco-tab-ablation} in Appendix~\ref{sec:ablation}, omitting semantic similarity from the kernel matrix drops the performance significantly. We use $S_{ij} = \text{IoU}_{ij} \times \text{sim}_{ij}^4$ for inferring object proposals in all our experiments.

\subsection{Smaller Number of Proposals}
To show the effectiveness of LDDP in capturing informative proposals, we restrict the number of proposals generated by LDDP to a fraction of those generated by Faster R-CNN on VOC2007 data set. Limiting the number of proposals generated by our LDDP model to 100 drops mean AP on VOC2007 test set from $62.2\%$ to $60.4\%$ which is similar to the mean AP achieved by 300 proposals in Faster R-CNN. One can refer to Appendix~\ref{sec:num_det}, Figure~\ref{fig:num_dets} for a complete plot representing performance versus number of proposals. This experiment approves high-confidence non-redundant localization through our end-to-end LDDP network.

\subsection{Visualization}
\begin{figure*}
\centering
\includegraphics[width=1\textwidth]{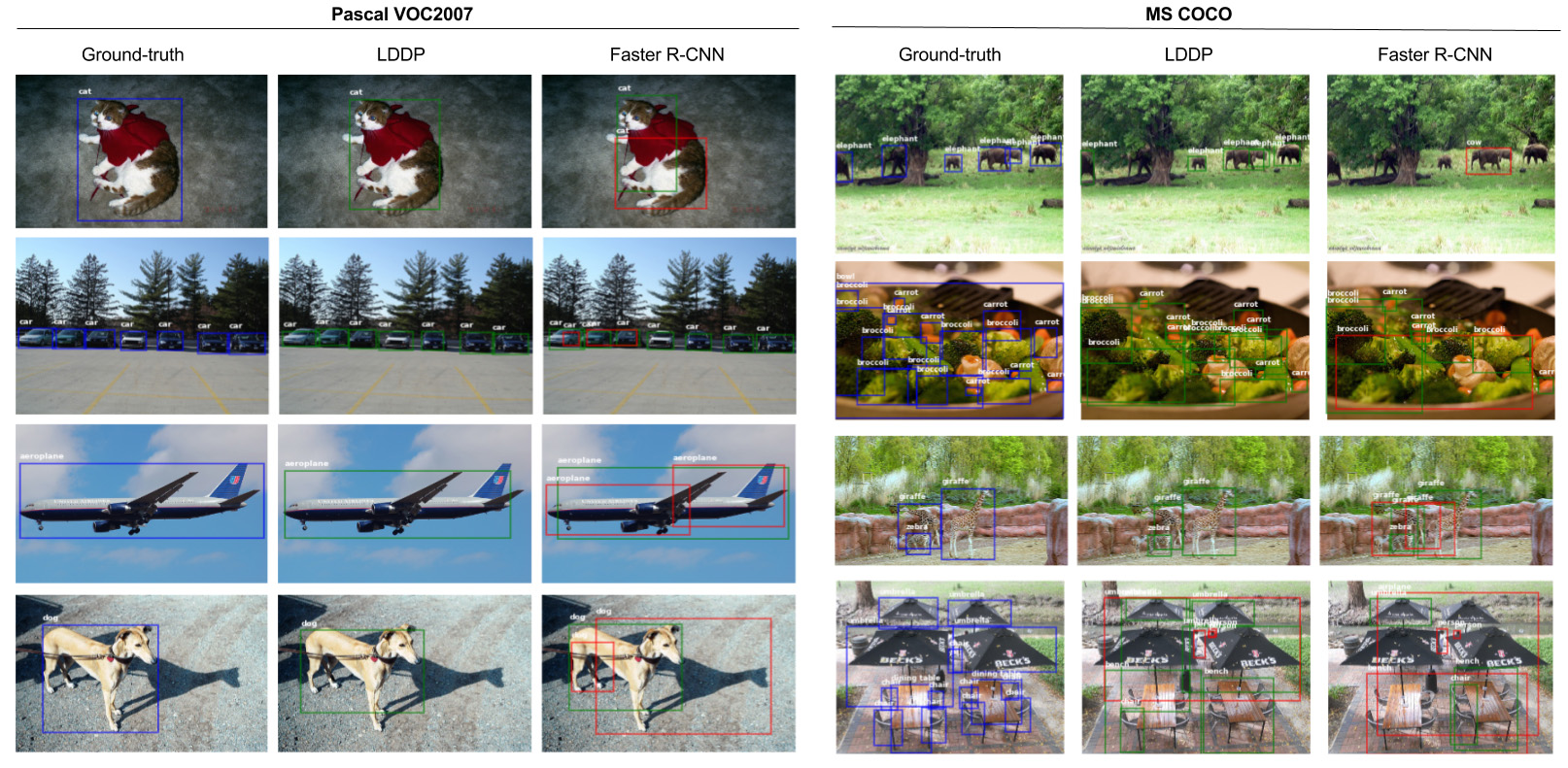}
\caption{Example images from Pascal VOC2007 and MS COCO data sets illustrating LDDP and Faster R-CNN networks used for learning. A score threshold of 0.6 is used to display images. NMS is used for pruning proposals.}
\label{fig:learning}
\end{figure*}

\begin{figure*}[!t]
\centering
\includegraphics[width=1\textwidth]{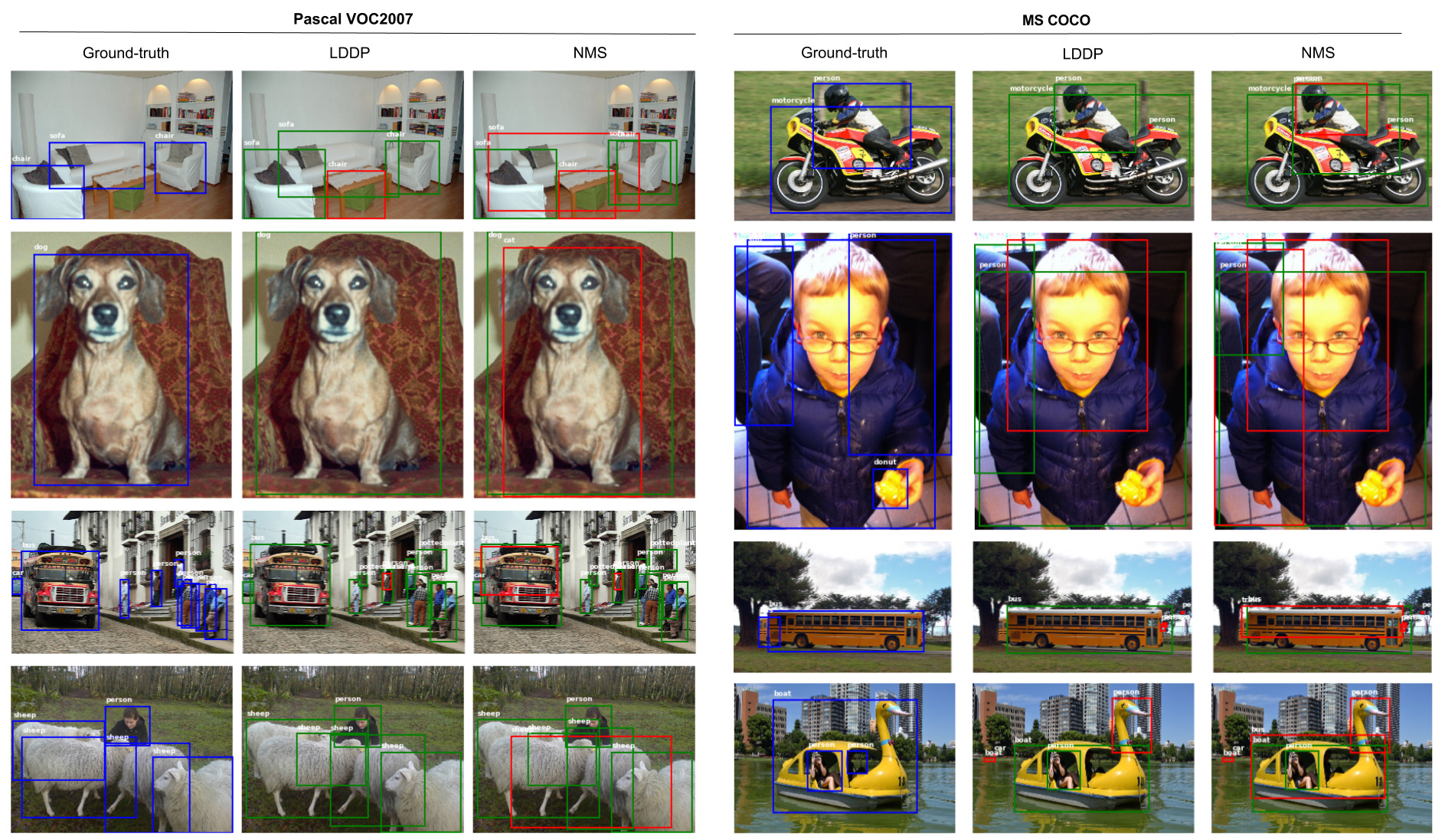}
\caption{Example images from Pascal VOC2007 and MS COCO data sets illustrating LDDP inference and NMS applied on top of Faster R-CNN predicted detections. A score threshold of 0.6 is used to display images.}
\label{fig:infer}
\end{figure*}

We visualize the output of both our learning and inference LDDP models in Figures~\ref{fig:learning},~\ref{fig:infer} in comparison with the state-of-the-art detection model, Faster R-CNN, and NMS for pruning the proposals. Specifically, we illustrate the predictions of our proposed LDDP model as well as the baseline for the learning step on both Pascal VOC2007 and MS COCO data sets for prediction scores above 0.6 in Figure~\ref{fig:learning}. We use NMS for the inference step in both predictions. Similarly, we compare the pruned detection boxes by the LDDP and NMS algorithms in Figure~\ref{fig:infer} using the Faster R-CNN training model. Bounding boxes found by our LDDP network are both more diverse and representative of the image content as illustrated.

\section{Conclusion}
We presented a deep network architecture for object detection which considers all contextual information, spatial layout, and label-level relationships between proposals during training and inference procedures and reinforces the parameters of the model to move toward a more accurate localization of bounding boxes and a better representation. The experimental results on PASCAL VOC and MS COCO benchmark data sets demonstrate the superiority of the LDDP detection performance in terms of average precision and recall. It is worth noting that our model does not add any new parameters to the deep network while boosting the performance and also results in the same detection performance as Faster R-CNN only by generating 1/3 proposals.

Learning semantic weights~\cite{desai2011discriminative} instead of using a fixed semantic similarity matrix based on WordNet hierarchy can be investigated as a future study. As another possible future work, applying our DPP loss layer in the region proposal network besides the multi-task loss in the detection network can also reinforce more efficiency among generated proposals and boost the detection performance. 

\vspace{0.3cm}
\textbf{Acknowledgement} The authors would like to thank Marcus Rohrbach and Fisher Yu for helpful discussions. This work was supported in part by DARPA; NSF awards  IIS-1212798, IIS-1427425, and IIS-1536003, Berkeley DeepDrive, and the Berkeley Artificial Intelligence Research Center.

\appendix
\section{LDDP Back-Propagation}
\label{sec:proof}
Given the set of representative proposals, $Y$, and set of probable background proposals, $B$, we can compute the log-likelihood loss function in Eq.~\eqref{log-like-s} as well as its gradient with respect to the confidence scores.
\begin{eqnarray}
\mathcal{L}(\alpha) =&&  \log P_{\alpha}(Y|X) -  \log P_{\alpha}(B|X) \label{log-like-s}
\end{eqnarray} 
where $\alpha$ refers to the parameters of the deep network, and conditional probabilities are defined based on the DPP model~\cite{kulesza2012determinantal} and our set of parameters:
\begin{eqnarray}
\label{DPP-s}
P_{\alpha}(\mathbf{Y} = Y|X) &=& \frac{1}{\det(L + I)} \det L_Y, \nonumber  \\
L_{i,j} &=& \Phi_i^{1/2}S_{ij}\Phi_j^{1/2}.
\end{eqnarray}
where $L$ denotes the L-ensemble matrix, $S$ the similarity matrix, $\Phi$ the quality measure, and $L_{Y} := [L_{ij}]_{i,j\in {Y}}$ denotes the restriction of $L$ to the entries indexed by elements of ${Y}$. Expanding the above probability distribution:
\begin{eqnarray}
\label{logp-s}
P_{\alpha}(Y|X) &=& \bigg( \prod_{i\in Y} \Phi_i \bigg) \frac{\det S_{Y}}{\det (L+I)},\nonumber\\
\log P_{\alpha}(Y|X) &=& \sum_{i \in Y} \log \Phi_i \nonumber\\
&+& \log \det S_{Y} -\log \det (L+I)
\end{eqnarray}
where $\det (L+I)$ is the normalizing factor as $\sum_{Y'\subseteq \mathcal{Y}} L_{Y'}$ with $\mathcal{Y}$ as all possible sets of proposals selections.

Now, we take the gradient of each term of the above loss function with respect to the outputs of the inner product layer before softmax. As explained in the paper, for the first term, $p_1=\log P_{\alpha}(Y|X)$, we have:
\begin{eqnarray}
\Phi_{i} = \begin{cases} 
\text{IoU}_{i,gt^i} \times \exp\{W_{gt}^Tf_i\} ,& \text{if } i \in Y\\
 \text{IoU}_{i,gt^i} \times \sum_{c \neq 0}\exp\{W_{c}^Tf_i\} &\text{if } i \not \in Y
\end{cases}
\label{phi-p-s}
\end{eqnarray}
where $W_{gt}$ denotes the weight vector for the corresponding ground-truth label of proposal $i$, and $c=0$ shows the background category. According to Eq.~\eqref{logp-s},~\eqref{phi-p-s}, the first conditional probability distribution would be as:
\begin{eqnarray}
\label{log-py-s}
\log P_{\alpha}(Y|X) &=& \sum_{i \in Y} \log \text{IoU}_{i,gt^i}  + \sum_{i \in Y}b_i^{gt}\nonumber \\&+& \log \det S_{Y} -\log \det (L+I)
\end{eqnarray}
where $b_i^{gt}=W_{gt}^Tf_i$. The proposals indexed by $i \not \in Y$ and labeled as background are not involved in this log probability resulting in a zero gradient. The same result will be applied on the proposals indexed by $i \in Y$ and labeled by category $c\neq gt$. On the other hand:
\begin{eqnarray}
\label{normalization-s}
\log \det (L+I) = \log \sum_{Y'} \bigg(\prod_{j \in Y'} \Phi_j\bigg) \det S_{Y'}
\end{eqnarray}

\begin{table*}[!t]
\centering
\caption{Ablation study on semantic similarity matrix used in LDDP inference. MS COCO minival detection average precision and average recall($\%$) (trained on COCO train set). All methods use \text{VGG\_CNN\_M\_1024} deep convolutional network.}
\label{coco-tab-ablation}
\begin{tabular}{|l|lll|lll|lll|lll|}
\hline
\multirow{2}{*}{Similarity Matrix} & \multicolumn{3}{l|}{Avg Precision @ IoU:} & \multicolumn{3}{l|}{Avg Precision @ Area:} & \multicolumn{3}{l|}{Avg Recall, \#Dets:} & \multicolumn{3}{l|}{Avg Recall @ Area:} \\ 
                        & 0.5-0.95      & 0.5         & 0.75        & S            & M            & L            & 1            & 10          & 100         & S           & M           & L           \\ \hline
$S_{ij} = \text{IoU}_{ij} \times \text{sim}_{ij}$      & 15.4       & 32.0      & 13.0      & 4.0       & 16.3       & 25.0      & 17.1       & 24.9      & 25.4     &   7.1    & 27.5      & 41.6      \\ 
$S_{ij} = \text{IoU}_{ij} \times \text{sim}_{ij}^4$            & 15.4        & 32.3      & 13.1     & 4.0        & 16.5       & 25.2       & 17.4       & 25.6      & 26.1      & 7.5      & 28.4      & 42.9      \\
$S_{ij} = \text{IoU}_{ij}$             & 14.5        & 29.9     & 12.5      & 3.6        & 15.3    &23.6   & 15.4      & 21.5       & 21.9      & 5.7      & 23.3      & 35.1     \\ 
 \hline
\end{tabular}
\end{table*}
Therefore, according to Eq.~\eqref{phi-p-s},~\eqref{normalization-s} for the proposals indexed by $i \in Y$ and labeled $c=gt$:
\begin{eqnarray}
\label{diff_L-s}
&&\frac{\partial\log \det (L+I) }{\partial b_i^c} \nonumber\\
&&=\sum_{Y'}\text{I}\{i \in Y'\} \frac{\partial \Phi_i}{\partial b_i^c} \bigg(\prod_{\substack{j \in Y'\\ j \neq i}} \Phi_j\bigg) \frac{\det S_{Y'}}{ \det (L+I)} \\
&&=\sum_{Y'}\text{I}\{i \in Y'\} \bigg( \prod_{j \in Y'} \Phi_j \bigg) \frac{ \det S_{Y'}} {\det (L+I)} = K_{ii}\nonumber 
\end{eqnarray}
Here, $K_{ii} = L_{ii}/\det(L+I)$, and $\text{I}\{.\}$ is the indicator function. Combining Eq.~\eqref{log-py-s},~\eqref{diff_L-s}:
\begin{eqnarray}
\label{logp-i}
\frac{\partial log p_1}{\partial b_i^c} = 1-K_{ii} \quad \forall i \in Y, c=gt
\end{eqnarray}
Similarly for the proposals indexed by $i \not \in Y$ and labeled as $c\neq 0$:
\begin{eqnarray}
\label{diff_Lp-s}
&&\frac{\partial\log \det (L+I) }{\partial b_i^c} \nonumber\\
&&=\sum_{Y'}\text{I}\{i \in Y'\} \frac{\partial \Phi_i}{\partial b_i^c} \bigg( \prod_{\substack{j \in Y'\\ j \neq i}} \Phi_j\bigg) \frac{\det S_{Y'}}{ \det (L+I)}\nonumber \\
&&=\sum_{Y'}\text{I}\{i \in Y'\} \frac{\exp \{b_i^c\}}{\sum_{c'\neq 0}\exp\{b_i^{c'}\}} \bigg(\prod_{j \in Y'} \Phi_j\bigg) \frac{\det S_{Y'}}{\det (L+I)}\nonumber \\
&&= K_{ii} \frac{\exp \{b_i^c\}}{\sum_{c'\neq 0}\exp\{b_i^{c'}\}}
\end{eqnarray}
Again, using Eq.~\eqref{log-py-s},~\eqref{diff_Lp-s} results in:
\begin{eqnarray}
\label{logp-ip-s}
\frac{\partial log p_1}{\partial b_i^c} = -K_{ii} \frac{\exp \{b_i^c\}}{\sum_{c'\neq 0}\exp\{b_i^{c'}\}} \quad \text{if } i \not \in Y, c\neq 0
\end{eqnarray}
Thus based on Eq.~\eqref{logp-i},~\eqref{logp-ip-s}, the gradient of the first log likelihood can be summarized as:

\begin{eqnarray}
\label{grad-p-s}
\frac{\partial \log p_1}{\partial b_i^c} = \begin{cases} 
1-K_{ii} ,& \text{if } i \in Y, c=\text{gt}\\
 \frac{ -K_{ii} \exp\{b_i^c\}}{\sum_{c'\neq 0} \exp\{b_i^{c'}\}}, &\text{if } i \not \in Y,  c\neq 0\\
0 & \text{otherwise}
\end{cases}
\end{eqnarray}

For the second log probability, $\log p_2 =\log P_{\alpha}(B|X)$, we change the quality measures as discussed in the paper. We skip the derivation of gradient of $\log p_2$ with respect to each $b_i^c$, which can be achieved by following a similar scheme:
\begin{eqnarray}
\label{grad-q-s}
\frac{\partial \log p_2}{\partial b_i^c} = \begin{cases} 
-K_{ii} ,& \text{if } i \not \in B,  c=\text{gt}\\
 \frac{-(K_{ii}-1) \exp\{b_i^c\} }{\sum_{c'\neq 0} \exp\{b_i^{c'}\}}, &\text{if } i \in B,  c\neq 0\\
0 & \text{otherwise}
\end{cases}
\end{eqnarray}

\section{Additional Experiments}

\subsection{Smaller Number of Proposals}
\label{sec:num_det}
\begin{figure}[t]
\centering
\includegraphics[width=0.5\textwidth]{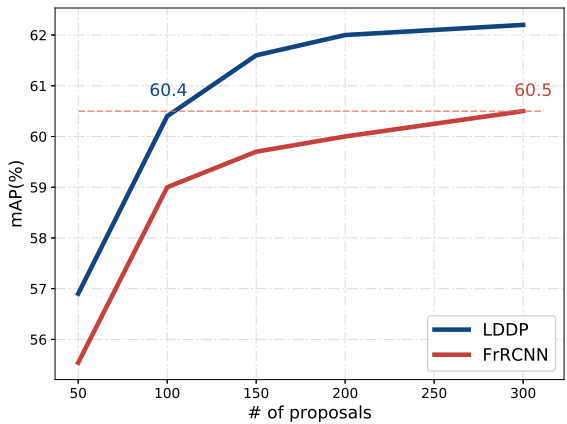}
\caption{Detection mAP($\%$) vs. number of proposals generated by our end-to-end LDDP model and Faster R-CNN. Both methods use ZF deep convolutional network and are trained on VOC2007 trainval.}
\label{fig:num_dets}
\end{figure}

As explained in the paper, to approve the generation of high-confidence non-redundant proposals through our proposed LDDP network, we evaluate bounding box detection performance when we restrict the number of generated proposals to different values, as shown in Figure~\ref{fig:num_dets}. Limiting the number of proposals generated by our LDDP model to 100 drops mean AP on VOC2007 test set from $62.2\%$ to $60.4\%$ which is similar to the mean AP achieved by 300 proposals in Faster R-CNN network ($60.5\%)$. Thus, our LDDP model is much more efficient than the state-of-the-art Faster R-CNN approach for the task of object detection.

\subsection{Ablation Study on Microsoft COCO}
\label{sec:ablation}
To understand how the semantic similarity matrix used in the kernel matrix $L$ affects the performance of our LDDP model, we use its different powers during inference and evaluate the detection performance on the minival subset of MS COCO data set with 5K images. According to the results reported in Table~\ref{coco-tab-ablation}, the semantic similarity matrix plays a crucial role in achieving accurate boxes.

\subsection{Visualization}
\begin{figure*}
\centering
\includegraphics[width=0.96\textwidth]{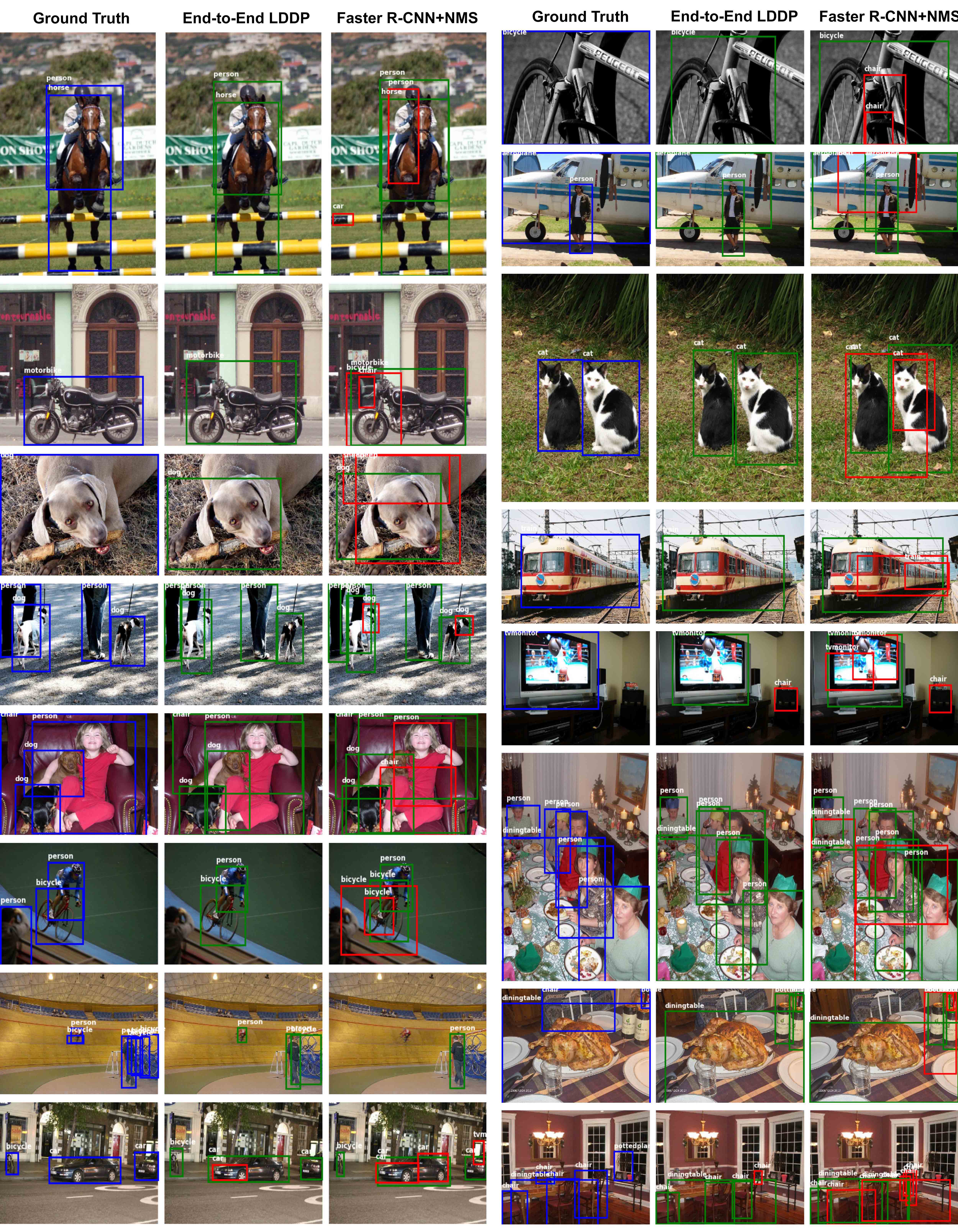}
\caption{Example images from Pascal VOC2007 data set illustrating our end-to-end LDDP and Faster R-CNN followed by NMS. A score threshold of 0.6 is used to display images. 
}
\label{fig:pascal}
\end{figure*}

\begin{figure*}
\centering
\includegraphics[width=0.96\textwidth]{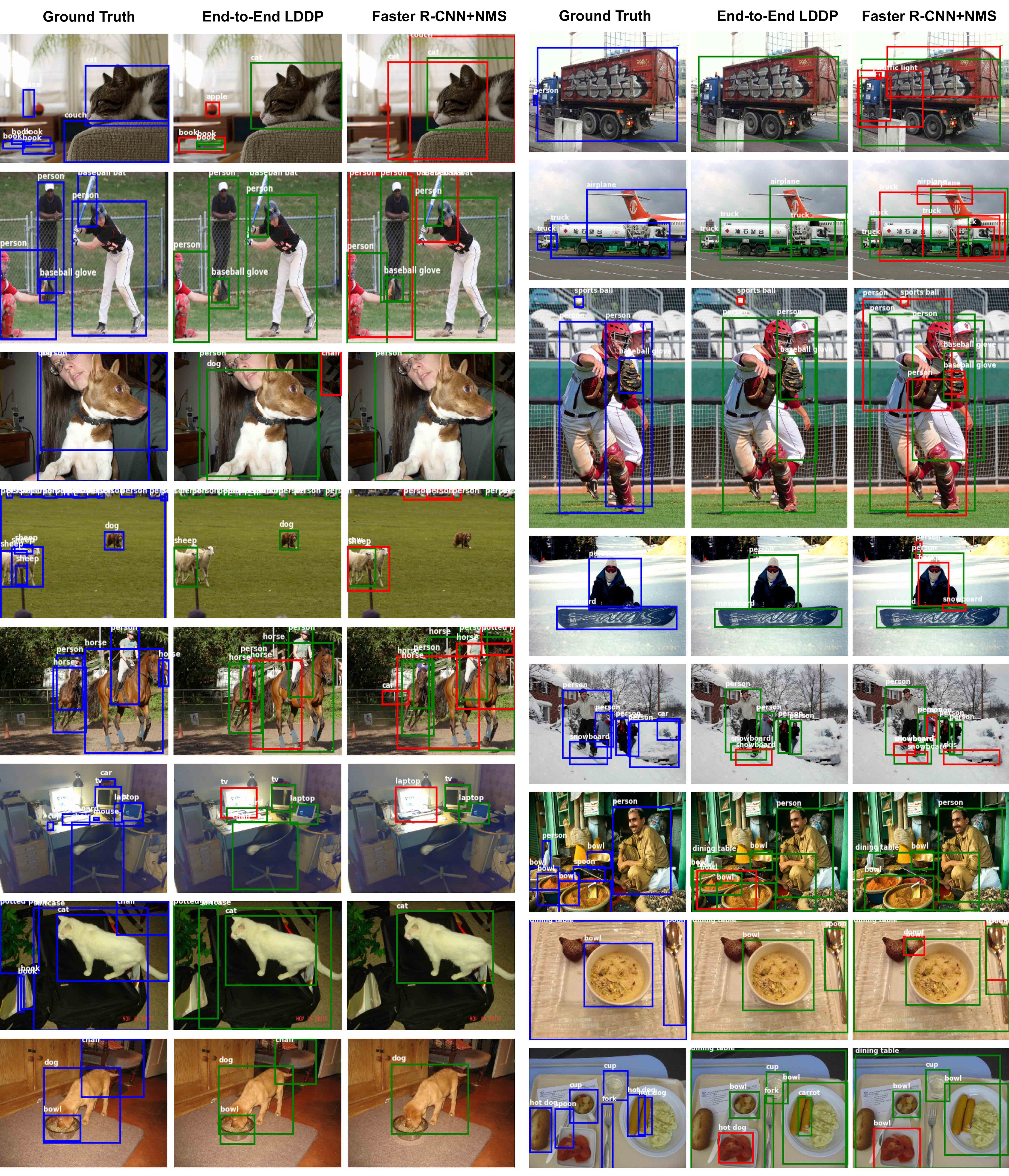}
\caption{Example images from MS COCO data set illustrating our end-to-end LDDP and Faster R-CNN followed by NMS. A score threshold of 0.6 is used to display images. 
}
\label{fig:coco}
\end{figure*}

We visualize the output of our end-to-end LDDP model as well as Faster R-CNN followed by NMS both on Pascal VOC2007 and MS COCO data sets~\cite{lin2014microsoft} in Figures~\ref{fig:pascal} and ~\ref{fig:coco}, respectively. We use the ZF model architecture for training the models on Pascal VOC2007 data set and the \text{VGG\_CNN\_M\_1024} deep network for training on MS COCO. The non-repetitive and accurate detections by the LDDP model reveal the superiority of our model against Faster R-CNN.

{\small
\bibliographystyle{ieee}
\bibliography{newBib}
}
\end{document}